\documentclass[a4paper]{article}

\usepackage{arxiv}

\usepackage[utf8]{inputenc} 
\usepackage[T1]{fontenc}    
\usepackage{hyperref}       
\usepackage{url}            
\usepackage{booktabs}       
\usepackage{amsfonts}       
\usepackage{nicefrac}       
\usepackage{microtype}      
\usepackage{cleveref}       
\usepackage{graphicx}
\usepackage{natbib}
\usepackage{doi}

\usepackage{subcaption}     
\usepackage[resetlabels,labeled]{multibib}  

\newcites{Apx}{References (Appendix)}

\title{Dynamic Personality Adaptation in \\Large Language Models via State Machines}

\newif\ifuniqueAffiliation

\ifuniqueAffiliation 
\author{ \href{https://orcid.org/0000-0000-0000-0000}{\includegraphics[scale=0.06]{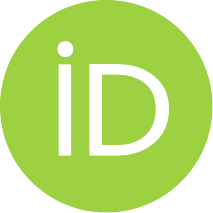}\hspace{1mm}David S.~Hippocampus}\thanks{Use footnote for providing further
		information about author (webpage, alternative
		address)---\emph{not} for acknowledging funding agencies.} \\
	Department of Computer Science\\
	Cranberry-Lemon University\\
	Pittsburgh, PA 15213 \\
	\texttt{hippo@cs.cranberry-lemon.edu} \\
	\And
	\href{https://orcid.org/0000-0000-0000-0000}{\includegraphics[scale=0.06]{orcid.pdf}\hspace{1mm}Elias D.~Striatum} \\
	Department of Electrical Engineering\\
	Mount-Sheikh University\\
	Santa Narimana, Levand \\
	\texttt{stariate@ee.mount-sheikh.edu} \\
}
\else
\usepackage{authblk}

\newbox{\orcid}\sbox{\orcid}{\includegraphics[scale=0.06]{orcid.pdf}} 
\author[1,2]{%
	\href{https://orcid.org/0009-0003-0933-6424}{\usebox{\orcid}\hspace{1mm}Leon Pielage}%
}
\author[3]{%
	\href{https://orcid.org/0009-0009-1410-4023}{\usebox{\orcid}\hspace{1mm}Ole Hätscher}%
}
\author[3]{%
	\href{https://orcid.org/0000-0003-2186-1558}{\usebox{\orcid}\hspace{1mm}Prof. Dr. Mitja Back}
    
}
\author[4]{%
	\href{https://orcid.org/0000-0002-1354-8687}{\usebox{\orcid}\hspace{1mm}Prof. Dr. med. Bernhard Marschall}%
}
\author[1,2]{%
	\href{https://orcid.org/0000-0001-5691-4029}{\usebox{\orcid}\hspace{1mm}Prof. Dr. Benjamin Risse\thanks{\texttt{b.risse@uni-muenster.de}}}%
}

\affil[1]{Institute for Geoinformatics, University of Münster, 48149 Münster, Germany}
\affil[2]{Faculty of Mathematics and Computer Science, University of Münster, 48149 Münster, Germany}
\affil[3]{Department of Psychology, University of Münster, 48149 Münster, Germany}
\affil[4]{Institute of Medical Education and Student Affairs, University of Münster, 48149 Münster, Germany}
\fi

\renewcommand{\headeright}{Preprint}
\renewcommand{\undertitle}{Preprint}
\renewcommand{\shorttitle}{Dynamic Personality Adaptation in LLMs via State Machines}

\hypersetup{
pdftitle={Dynamic Personality Adaptation in Large Language Models via State Machines},
pdfsubject={cs.HC},
pdfauthor={Leon Pielage, Ole Hätscher, Prof. Dr. Mitja Back, Prof. Dr. med. Bernhard Marschall, Prof. Dr. Benjamin Risse},
pdfkeywords={LLM, affective computing, personality, HCI, IPC},
}

\begin{document}
\maketitle              
\begin{abstract}
The inability of Large Language Models (LLMs) to modulate their personality expression in response to evolving dialogue dynamics hinders their performance in complex, interactive contexts.
We propose a model-agnostic framework for dynamic personality simulation that employs state machines to represent latent personality states, where transition probabilities are dynamically adapted to the conversational context.
Part of our architecture is a modular pipeline for continuous personality scoring that evaluates dialogues along latent axes while remaining agnostic to the specific personality models, their dimensions, transition mechanisms, or LLMs used. 
These scores function as dynamic state variables that systematically reconfigure the system prompt, steering behavioral alignment throughout the interaction.
We evaluate this framework by operationalizing the Interpersonal Circumplex (IPC) in a medical education setting. 
Results demonstrate that the system successfully adapts its personality state to user inputs, but also influences user behavior, thereby facilitating de-escalation training. 
Notably, the scoring pipeline maintains comparable precision even when utilizing lightweight, fine-tuned classifiers instead of large-scale LLMs. 
This work demonstrates the feasibility of modular, personality-adaptive architectures for education, customer support, and broader human-computer interaction.

\end{abstract}


\section{Introduction}

Large Language Models (LLMs) are increasingly integrated into daily life, both in professional contexts and during leisure activities, offering assistance across a growing range of cognitive and communicative tasks~\cite{openai_introducing_2025,qwen_qwen25_2025}. 
Alongside advances in their technical capabilities, recent years have witnessed growing interest not only in what LLMs communicate, but also in how they do so. 
A key development in this regard is the introduction of customizable assistant "personas" in top-tier models, which allow users to tailor the communicative style of the assistant for a more personalized and coherent interaction experience.

Concurrently, researchers have begun to explore the capacity of LLMs to simulate and assess human personality. 
Several studies have shown that LLMs can reliably express personality differences and generate psychologically plausible responses across diverse scenarios~\cite{li_big5-chat_2025,vu_psychadapter_2025-1,wen_self-assessment_2024}. 
However, both model-level customization and existing research efforts have largely focused on static personality representations, capturing how an assistant behaves on average rather than how it adapts to the interpersonal dynamics of an unfolding conversation. 
Yet, in human-human interactions, one of the defining features of personality expression is its dynamic nature, as individuals adjust their communicative behavior in response to their interaction partner’s tone, behavior, and perceived personality~\cite{back_chapter_2021,kuper_dynamics_2021,leary_interpersonal_1957}. 
Such dynamic adaptation is essential for creating more realistic and engaging simulations in applied domains such as medical education, human resources, or gaming. 

While state-of-the-art LLMs exhibit some degree of user-contingent behavior, they remain limited in three important ways:
\begin{enumerate}
    \item Responses are often shaped by fixed defaults (such as persistent friendliness) regardless of user behavior;
    \item the internal mechanisms underlying adaptation are opaque and not directly modifiable; and
    \item there is currently no systematic or modular way to analyze, manipulate, or test dynamic personality adaptation in LLM-based agents.
\end{enumerate}

To address these limitations, we propose a modular framework for dynamic personality adaptation in LLMs that enables real-time adjustment of the model’s personality in response to the evolving interaction.
The framework consists of a state machine architecture in which each state encodes a specific momentary personality expression. 
Transition probabilities are continuously updated according to the linguistic content of incoming user messages.
These messages are processed by auxiliary classifiers that estimate relevant personality-dimension scores, which in turn update the assistant’s system prompt, guiding its subsequent communicative behavior.
By implementing dynamic personality adjustment through prompt-level modulation, the framework remains model-agnostic and forward-compatible, allowing seamless integration with current and future LLMs.

\section{Related Work}

Recent studies show that large language models can simulate a variety of personality constructs (Big~Five~\cite{li_big5-chat_2025}, Interpersonal Circumplex~\cite{vu_psychadapter_2025-1},  Dark~Triad~\cite{yang_pdf_2025}, etc.) but most work concentrates on static, dimensional Big~Five representations~\cite{li_big5-chat_2025}.
Two main methodological families have emerged:
\begin{enumerate}
    \item \textbf{Editing the model:} fine‑tuning, adapters, low-rank adaptation (LoRA), or mixture‑of‑experts approaches that embed stable personality representations directly into the model’s weights~\cite{vu_psychadapter_2025-1,wen_self-assessment_2024}.
    \item \textbf{Inducing personality via prompts:} supplying trait descriptors, prototypical adjectives, or psychometric items (often with numerical anchors) to steer the model’s output without altering its architecture~\cite{caron_identifying_2022,cho_scaling_2025}. 
    More sophisticated prompting leverages embedding‑level knowledge of trait organization or LLM‑enhanced prompts that amplify personality expression~\cite{yang_pdf_2025}.
\end{enumerate}

Evaluations report good validity, reliability, and internal coherence of the generated personalities~\cite{serapio-garcia_personality_2025,wang_evaluating_2025}, yet little attention has been paid to dynamic aspects of personality modeling, namely how personality expression may change over time in response to the interaction context. 

Personality inference from language is well-established via zero-shot prompting~\cite{derner_can_2024,sikstrom_personality_2025}, fine-tuning~\cite{shen_less_2025-1,wang_emotion_2024}, and supervised learning~\cite{alsini_using_2024}. 
Zero-shot and prompt-based methods now achieve accuracies partly comparable to supervised models~\cite{sikstrom_personality_2025}, with human evaluations showing high correspondence between self-rated and LLM-predicted traits, despite a potential positivity bias~\cite{derner_can_2024}. 
While these methods demonstrate strong internal consistency ($\alpha > .70$) and convergent validity~\cite{petrov_limited_2024}, they largely focus on static personality in text-rich or self-attributed data~\cite{sikstrom_personality_2025}. 
In contrast, dynamic interaction requires inferring momentary personality from sparse dialogue, an area that remains underexplored.

Recently, interest in dynamic approaches to personality has begun to grow, both in terms of generation and assessment.
For example, some studies explore how LLMs adjust their behavior in real-time within interactive and strategic settings, such as social dilemma games like the Prisoner's Dilemma~\cite{zeng_dynamic_2025}. 
Others focus on tracking interaction dynamics—i.e., the dyadic and temporal patterns that unfold across exchanges, through the alignment of textual content in both human–human~\cite{fischer_personality_2024} and AI–AI interactions~\cite{frisch_llm_2024,li_evolving_2024}.

Despite this growing interest, only a few studies draw upon established models in personality science, particularly interpersonal and dynamic models, which provide a robust framework for modeling such phenomena (e.g.,~\cite{back_chapter_2021,kuper_dynamics_2021}).
Notable exceptions include: Wang et al.~\cite{wang_emotion_2024}, who developed a personality recognition model based on personality dynamics theory that tracks and updates the user’s personality state throughout the interaction~\cite{nezhad_adaptive_2025}, who proposed a dynamic personality recognition and expression system for persuasive communication, modulating the influence of default and situational personality states; and ~\cite{tang_personafuse_2025-1}, who fine-tuned a model to express personality in a context-sensitive manner.

\section{Method}

\begin{figure}[b]
    \centering
    \includegraphics[width=.95\linewidth]{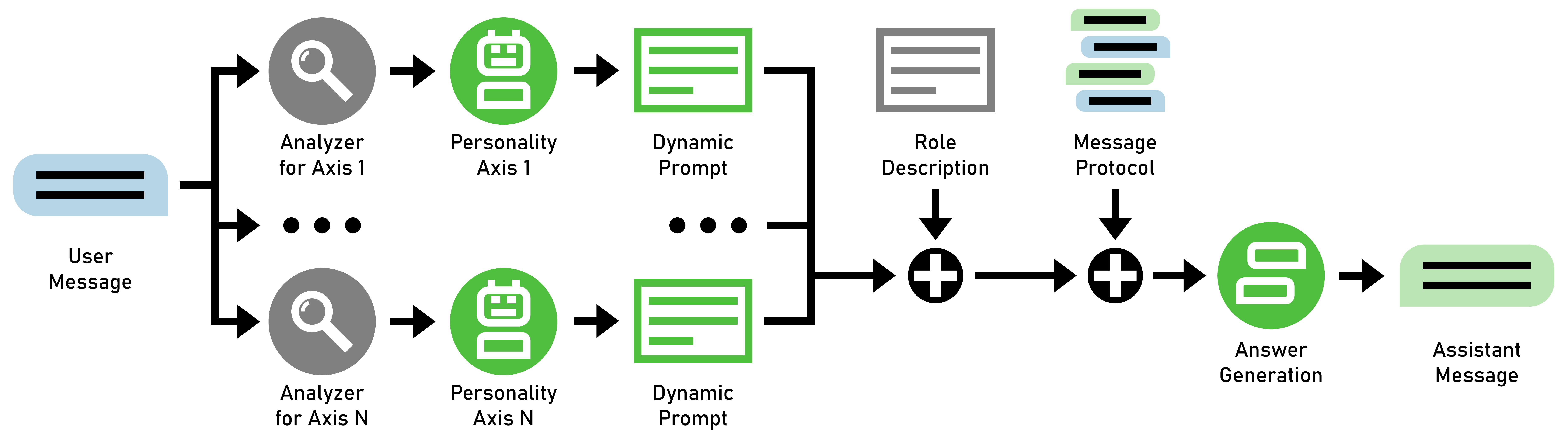}
    \caption{The analysis of incoming messages and the selection of a personality prompt happen per dimension. 
    Together with the role description and the message protocol, these provide the LLM-context.
    }
    \label{fig:full-pipeline}
\end{figure}

Our framework consists of two linked modules: an analyzer and a generative personality model. 
The analyzer predicts momentary personality states from user and assistant messages to determine how the generative model modulates the assistant’s communicative style along predefined personality dimensions. 
These dimensions, or axes, represent latent traits such as dominance/agency or warmth/communion. 
The framework supports an arbitrary number of axes, generally maintaining a one-to-one correspondence between the analyzer and generative model, to remain adaptable to various psychological theories.

Currently, we model each axis independently, implying that traits on different axes are treated as statistically independent factors.
This assumption can be relaxed by introducing one-to-many mappings where a personality state predicted by the analyzer influences multiple generative personality axes, or by cross-linking models to account for interdependencies between traits. 
Both modules can be implemented with models tailored to their respective functions, for instance a classifier for the analyzer and a generative model for response production. 
Our experiments employ dedicated LLMs prompted for their specific roles. 
This model-agnostic design allows underlying language models to be easily replaced as newer systems emerge.
The workflow including analysis, personality-state updates, and response generation is shown in \autoref{fig:full-pipeline}. 
In addition to a standard chat interface, a development mode provides helpful visualizations. 
In this mode, analyzed texts are highlighted and a separate plot displays the current personality state and transition probabilities, which facilitates inspection when personality inference from a single message is uncertain.

\subsection{Personality Dimensions}

The Interpersonal Circumplex (IPC)~\cite{gurtman_exploring_2009,leary_interpersonal_1957} describes behavior along two orthogonal dimensions: dominance (agency), reflecting assertiveness and control, and warmth (communion), capturing empathy and friendliness. 
These dimensions define a circular space for locating and interpreting interpersonal styles.
In our framework, these dimensions serve as examples for the personality axes used by the analyzer and generative model. 
However, our approach is not limited to the IPC and can extend to any personality model or number of dimensions required by the application.

\subsection{Message Analyzer}
\label{sec:analysis}

The analyzer maps messages to numeric scores along defined personality dimensions. 
This mapping can occur at the sentence or message level; however, our preliminary comparisons showed that message-level analysis is more effective because it leverages broader context and captures diagnostic cues distributed unevenly across sentences. 
Consequently, all subsequent experiments utilize message-level analysis.
We implemented three score-mapping strategies. 
First, off-the-shelf LLMs are prompted to output scores directly. 
Second, an LLM is specialized for the task via LoRA fine-tuning. 
Third, the model's final layer is replaced with a dedicated classification head to transform input representations into numeric scores. 
The following sections detail these strategies and the associated data generation process for the fine-tuned alternatives.

\subsubsection{Prompt Based Personality Mapping}
\label{sec:analysis:prompt}

A system prompt is formulated for each personality axis to instruct an LLM to give a score within a certain range. 
Depending on the axis, the prompt may be very short. 
However, providing examples improves the quality and reduces the number of cases where the model does not adhere to the output format, resulting in an unparseable score, especially for really small models. 
For LLMs which support this feature we also define a prefix to guide the model to output a single number without additional text.
An illustrative prompt for the agency dimension of the IPC model could be phrased as follows:
\begin{quote}
    \texttt{Give a score between 0 and 10 where 0 is very submissive and 10 is very dominant. 
    The score should be a single number.}
\end{quote}
In comparison a longer prompt gives more context about the personality model and might even contain an example:
\begin{sloppypar}
\begin{quote}
    \texttt{You are a helpful assistant that analyzes the dominance of a sentence according to the interpersonal circumplex model. 
    You will do this by giving a score between 0 and 10 where 0 is very submissive and 10 is very dominant. 
    The score should be a single number. 
    For example, 'No, you do it like this!' should be scored as 10.}
\end{quote}
\end{sloppypar}

\subsubsection{Data Generation}
\label{sec:analysis:data}

In order to fine-tune LLMs for the personality scoring task we generated a synthetic dataset using OpenAI GPT 4.1~\cite{openai_introducing_2025}. 
It consists of various sentences with labels for the two main axes of the IPC model both with a range from -5 to 5. 
1210 sentences have a label for both axes. 
Moreover, 500 additional sentences per axis have a label exclusively for dominance/agency or warmth/communion. 
The labels, like the sentences, were generated by GPT 4.1 by first generating a nuanced and standardized description per intensity step and then generating example sentences according to this description. 
While GPT 4.1 is able to generate realistic examples for all these intensities they might contain a bias and certainly do not cover all sentences frequently used on a chat conversation as GPT 4.1 tended to generate sentences which clearly show specific personality traits. 
In contrast, real conversations contain many utterances that are too short or, for other reasons, do not have a clear tendency on all personality axes, such as "Okay" or "No".
Therefore, we used 109 more realistic examples containing short and long messages from patients and doctors for the evaluation. 
These examples were human-validated and curated if necessary, and then rated by three human experts with a background in psychology.
We used the median to determine the target rating for comparison with our model predictions, as it is more robust to outliers than the mean. 
Both the training and test samples are in German.

\subsubsection{Fine-tuned Personality Mappings}
\label{sec:analysis:finetuning}

The above-mentioned dataset was used to fine-tune two small Qwen 2.5 models~\cite{qwen_qwen25_2025} with 0.5 billion parameters, applying two distinct fine-tuning strategies.
In the first strategy, we fine-tuned the LLM using the LoRA method~\cite{hu_lora_2021} to better adapt it to the scoring task. 
For each personality axis, a separate LoRA fine-tuning run was performed with a short prompt defining the corresponding dimension. 
We used a rank of 16, an alpha value of 32, and a dropout rate of 0.05 in all cases. The models were trained for 5 epochs with a batch size of 8 and a validation set size of 20\%. 
After training, we selected the best-performing model across epochs based on validation performance.
In the second strategy, we replaced the model’s final layer with a lightweight classification head that outputs the score directly. 
Using transfer learning, one model per personality axis was trained while keeping all other layers frozen. 
The batch size and validation set size were kept identical to the LoRA setup, while the number of training epochs was increased to 10. 
As before, the best model was selected according to validation performance.

\subsection{Personality Model}
\label{sec:personality}

\begin{figure}
    \centering
    \includegraphics[width=1.0\linewidth]{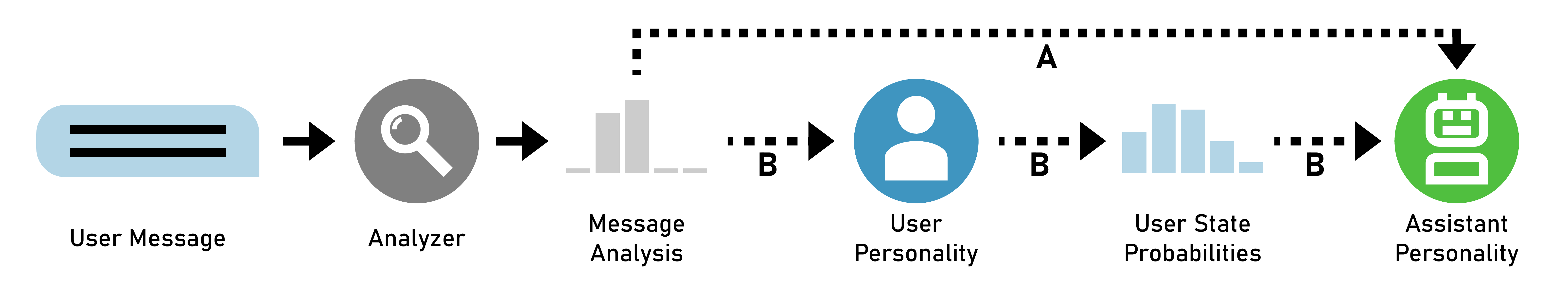}
    \caption{A user message is analyzed, resulting in scores according to a specific personality dimension. 
    These scores can \textbf{(A)} either be used directly to update the assistant personality or \textbf{(B)} be used to update a separate user personality model, which can then be used to update the assistant personality. 
    }
    \label{fig:analysis-update}
\end{figure}

The personality of the assistant is modeled as a state machine, in which each personality dimension is represented by a separate and independently operating axis. 
Each axis maintains its own set of states and transition probabilities, analyzing messages according to distinct criteria and, if desired, using different models. 
This modular design allows personality dimensions to vary in scope and implementation.
Axes can exist solely for the assistant or for both the assistant and the user. 
The latter configuration enables the system to track the user’s momentary personality throughout the interaction and to update the assistant’s transition probabilities based on the user’s current personality state rather than only on recent message content.
Each personality axis is influenced by multiple factors, most of which stem from internal representations and states that determine how transition probabilities evolve over time. 
At the same time, external influences allow the model to adapt dynamically to ongoing interactions. 
As described previously, these influences can be direct, based on the analysis results of the most recent message, or indirect, arising from the current personality state of another model (\autoref{fig:analysis-update}).

\subsubsection{State Machine}

The core of each personality axis is a state machine heavily influenced by probabilistic automata~\cite{rabin_probabilistic_1963} and Moore machines~\cite{church_edward_1958}.
We define it as $(S, \Sigma, \Gamma, \delta, \omega, s_d, F)$.
Following this definition we have a finite set of $k$ states $S = \{s_0, s_1, ..., s_{k-1} \}$ with a non-probabilistic starting state $s_d \in S$ and a set of accepting states $F=S$ as all our states are valid end states at the end of the input sequence.
An input sequence $(x_0, x_1, ..., x_n)$ of length $n$ consists of multiple elements from the input alphabet $\Sigma$.
Each input $x \in \Sigma$ consists of two discrete probability distributions over all states $x = (P_o, P_q) \in [0, 1]^{2\times k}$ which both sum to $1$.
Additional outside influences and the state probabilities are given by $P_o = (p(s_0), ..., p(s_{k-1})) \in [0, 1]^{k}$ and $P_q = (q(s_0), ..., q(s_{k-1})) \in [0, 1]^{k}$, respectively. 
The momentary personality expression depends on the current state $s \in S$ and is described in the output alphabet $\Gamma$ for each state.
Using the output function $\omega: S \rightarrow \Gamma$ it can be determined for each state.
In our case $\Gamma$ contains pre-defined system prompts for the LLM.
However, these can be replaced with other model inputs like parameter sets to control different downstream model types.
The state transition function is defined as $\delta: S \times \Sigma \rightarrow [0, 1]^k$ with $k = |S|$ being the number of states.
For a state $s \in S$ and an input $x \in \Sigma$ the transition probability going into state $s_i \in S$ is given through the probability $p_i(s, x)$.
Therefore, $\delta(s, x) = (p_0(s, x), ..., p_{k-1}(s, x))$ with $\sum_i p_i(s, x) = 1$.
Contrary to the definition of probabilistic automata~\cite{rabin_probabilistic_1963} our transition probabilities are not fixed but re-calculated in each iteration.

The next state is determined probabilistically based on these transition probabilities.
In practice we also implemented a deterministic mode in which the next state is not chosen based on the transition probabilities but always as the state with the highest probability, effectively make it to a deterministic state machine.
This greatly improves model stability and is applied to personality tracking for the user, where less exploratory behavior is desired.

\subsubsection{Transition Probabilities}

While our model allows transitions between any pair of states (including remaining in the same state), psychological theories of personality typically assume that transitions to neighboring states are more likely than transitions to distant ones~\cite{gurtman_exploring_2009}.
As each state machine models only one personality dimension, neighboring states can be defined by ordering them along that dimension.
Let the order of the states $s_0, ..., s_{k-1}$ be given through their index $0, ..., k-1$.
To approximate the transition probabilities to neighboring states we use multiple normal distributions.
The full transition probabilities are calculated in each iteration from four probability distributions:
\begin{itemize}
    \item \textbf{Default State} The default state is also the starting state $s_d$. 
    We use a normal distribution $\mathcal{N}(s_d; \sigma^2)$ to spread the probability of being in the region of the default state.
    \item \textbf{Current State} To achieve a stable personality simulation, the current state $s_c$ has a high influence of the transition probabilities for the next state. 
    Like the default state, it is approximated by a normal distribution $\mathcal{N}(s_c; \sigma^2)$.
    \item \textbf{Old State Probabilities} The old state probabilities also influence the next transition probabilities, introducing a certain temporal inertia into the interaction. 
    They are provided as part of the input $x \in \Sigma$.
    \item \textbf{Outside Influences} These allow the model to adapt dynamically throughout the interaction. 
    No specific distributional form is required, and we implemented two main sources of state probabilities: 
    (1) direct analyzer results for each message, where $p(s) = 1$ for a specific state $s$ if the entire message is analyzed as a single unit rather than split into sentences; and 
    (2) state probabilities $P'_q$ obtained from a different personality axis (typically the same personality dimension in another personality model, e.g., the assistant's communion axis is updated according to the state probabilities of the user's communion axis). 
    Both variants are illustrated in \autoref{fig:analysis-update}.
    To achieve a positive or negative correlation, the state probabilities can be mirrored beforehand by reversing the state order.
\end{itemize}

The new transition probabilities are calculated as a weighted linear combination of these four components whereas weights $w$ further modulate the dynamics of a respective personality:
\[
    \delta(s_c, (P_o, P_q)) = w_d * \mathcal{N}(s_d; \sigma^2) + w_c * \mathcal{N}(s_c; \sigma^2) + w_q * P_q + w_o * P_o
\]
for a state $s_c$, input $(P_o, P_q)$ and the default state $s_d$.
The state probabilities also need to be recalculated to be used as input for the next transition.
We calculate the new state probabilities $\hat P_q$ equivalent to the transition probabilities but without the influence of the current state:
\[
    \hat P_q = \hat w_d * \mathcal{N}(s_d; \sigma^2) + \hat w_q * P_q + \hat w_o * P_o.
\]
The weights $w_d$, $w_c$, $w_q$ and $w_o$ sum to one and can be used to influence how fast the assistant adapts its momentary personality expression to the interaction, stays in its current state or falls back to the default state.
To calculate the state probabilities, we rescale them to satisfy: $\hat w_d + \hat w_q + \hat w_o = 1$.
Additionally, the standard deviation $\sigma$ used for the normal distributions around the default and the current state controls how reactive the assistant is and how likely it will explore neighboring states.
Empirically, we found that assigning a high weight to the current state ($w_c = 0.5$) stabilizes the system, while keeping the influence of the default state small ($w_d = 0.1$).
For the assistant, we found that lower weights on the outside influence ($w_o = 0.1$) and a small reactivity ($\sigma = 0.1$) led to more natural interactions when the updates were based on the user’s personality.
At the same time, we used the deterministic mode for the user personality and therefore selected higher values ($w_o = 0.2$, $\sigma = 0.6$) to focus more on tracking the momentary personality based on each new message and to avoid exploratory behavior.

\subsubsection{Generation}

\begin{figure}
    \centering
    \includegraphics[width=1.0\linewidth]{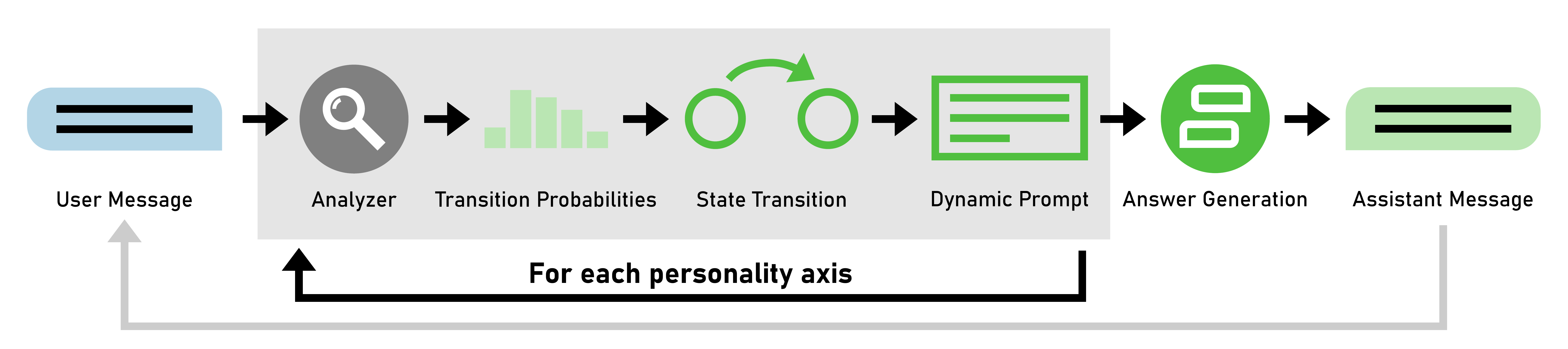}
    \caption{Steps to generate a reply to an incoming user message.}
    \label{fig:answer-generation}
\end{figure}

\autoref{fig:answer-generation} visualizes how the answer to a user message is generated.
Based on the analyzer (\autoref{sec:analysis}) results for a specific personality dimension the transition probabilities in the corresponding personality axis are updated and the next state is sampled.
This state provides a textual description of how a given level of this personality dimension is expressed.
The full system prompt for the LLM comprises two components (\autoref{fig:full-pipeline}). 
A fixed part that encodes static information about the interaction context, such as the name and age of a virtual agent (in our experiments acting as a patient), which we refer to as a role description.
The second component is the dynamic part that is continuously updated based on the textual descriptions from all personality axes.
Since they are independent of each other, they can be combined freely, thereby increasing the number of possible overall personality states of the assistant.

In our experiments we focused on the IPC model~\cite{leary_interpersonal_1957,gurtman_exploring_2009} with its two main dimensions agency (or dominance) and communion (or warmth).
We approximated these dimensions in two axes, each with five states. 
Each state has a prompt similar to the following, for a less agentic state:
\begin{quote}
    \texttt{You are rather cautious and prefer to leave decisions to the other person. You rarely express yourself proactively and only address complaints or wishes with restraint. When you communicate something, you phrase it as cautiously as possible, perhaps with sentences like "I'm not sure, but...". You are easily influenced, but polite and cooperative.}
\end{quote}
Due to the independence of the axes, the resulting personality model comprises 25 individual states.

\section{Evaluation}

We conducted two experiments to evaluate the feasibility of personality inference from short text inputs and the effectiveness of our adaptive personality in interactions. 
The first compared various LLMs on analyzing natural dialogue messages, focusing on the potential for improvement via fine-tuning. 
The second let participants interact with our prototype to qualitatively assess the modeling pipeline and the plausibility of personality adaption.
In both experiments, we used a configuration according to the IPC model with its two main dimensions agency and communion. 
All generated texts and user interactions were performed in German.

\subsection{Analyzer Comparison and Fine-tuning}
\label{sec:eval-analyzer}

Inferring personality from sparse linguistic contexts is a known challenge for LLMs. 
Initial tests suggested that even top-tier models like GPT-4.1 struggle to generate accurate ratings for short German texts. 
To systematically evaluate this, we benchmarked several models against a human-annotated dataset of 109 messages (\autoref{sec:analysis:data}). 
To ensure the quality of this evaluation, we calculated the Intraclass Correlation (ICC) between the three trained annotators, yielding values of 0.602 for Agency and 0.581 for Communion, indicating moderate to good inter-rater reliability for such a subjective task.

Our analyzer module can operate as either a prompted LLM or a dedicated classifier. 
While LLMs are well-suited for these tasks due to their deep linguistic understanding, we hypothesized that simple fine-tuning could significantly enhance the performance of very small models. 
We applied two strategies to the Qwen 2.5 0.5B model using synthetic training data (\autoref{sec:analysis:data}): a LoRA adapter to align the model with the scoring task and transfer learning with a new classification head.

We compared our results to our purely prompt based approach using the baseline Qwen 2.5 0.5B model~\cite{qwen_qwen25_2025} and other popular LLMs, namely Llama 3.3 70B, Gemma 3 27B, Mistral Small 3.1 24B and the GPT 4.1 model family.
For each model, we used both a short and a long system prompt for the analysis (\autoref{sec:analysis:prompt}). 
For each variant, we calculated both the accuracy and a one-off accuracy, since neighboring states might also be valid predictions for this task.
Additionally, we report the mean distance between the predicted and the target score as well as the error rate.
The error rate reflects how often smaller models failed to follow instructions, producing unparseable text instead of a numeric score. 

\begin{table}
    \centering
    \caption{
        Performance of selected LLMs on the task of rating messages in terms of their \textbf{Agency}/\textbf{Communion} prompted with a short (s)/long (l) prompt.
        Accuracy (Acc.)/One-off accuracy (1-off)/Mean distance (Mean)/Error rate (Error)
    }
    \label{tab:fine-tune}
    \begin{tabular}{lcccccccc}
        \toprule
        Model Name & \multicolumn{4}{c@{}}{Agency} & \multicolumn{4}{c@{}}{Communion} \\
        \cmidrule(l){2-5}
        \cmidrule(l){6-9}
         & Acc. $\uparrow$ & 1-off $\uparrow$ & Mean $\downarrow$ & Error $\downarrow$ & Acc. $\uparrow$ & 1-off $\uparrow$ & Mean $\downarrow$ & Error $\downarrow$ \\
        \midrule
        Our Classify 0.5B & 0.2294 & 0.5138 & 2.1743 & - & 0.1560 & 0.4220 & 2.0642 & - \\
        Our LoRA 0.5B (s) & 0.2569 & 0.6239 & 1.5596 & 0.0000 & 0.0734 & 0.2294 & 2.8073 & 0.0000 \\
        Qwen 2.5 0.5B (s) & 0.0196 & 0.2647 & 3.0000 & 0.0642 & 0.0408 & 0.0918 & 3.9592 & 0.1009 \\
        Qwen 2.5 0.5B (l) & 0.0000 & 0.2075 & 3.1887 & 0.0275 & 0.0120 & 0.0602 & 3.9398 & 0.2385 \\
        Llama 3.3 70B (s) & \textbf{0.2981} & \textbf{0.7308} & \textbf{1.1346} & 0.0459 & 0.2018 & 0.4220 & 1.9817 & 0.0000 \\
        Llama 3.3 70B (l) & 0.1296 & 0.5185 & 1.9630 & 0.0092 & 0.2294 & 0.5688 & 1.5229 & 0.0000 \\
        GPT 4.1 mini (s) & 0.1009 & 0.3303 & 2.7248 & 0.0000 & 0.2294 & 0.5229 & 1.4587 & 0.0000 \\
        GPT 4.1 mini (l) & 0.0734 & 0.2661 & 2.6055 & 0.0000 & 0.2936 & \textbf{0.7339} & \textbf{1.0275} & 0.0000 \\
        GPT 4.1 (s) & 0.0185 & 0.1481 & 3.7685 & 0.0092 & 0.1651 & 0.4495 & 1.7706 & 0.0000 \\
        GPT 4.1 (l) & 0.0550 & 0.2477 & 2.8349 & 0.0000 & \textbf{0.3211} & 0.6881 & 1.0917 & 0.0000 \\
        \bottomrule
    \end{tabular}
\end{table}

Results (\autoref{tab:fine-tune}) show that fine-tuning significantly improves performance over non-fine-tuned baselines of the same model. 
Beyond raw accuracy, fine-tuning effectively eliminated the error rate, which is arguably more important in a real-time setting like ours.
For the agency dimension (\autoref{tab:fine-tune}), our LoRA-tuned model achieved 25.69\% accuracy, surpassed only by the Llama 3.3 70B model with 29.81\%.
Notably, many newer models perform worse than the older Llama 3.3 70B on this personality dimension, and performance decreases with a longer prompt for most models except GPT 4.1.
This aspect requires further investigation, since prompt formulation strongly affects performance and additional language-specific factors may also contribute.

In the communion condition (\autoref{tab:fine-tune}), the transfer learning model outperformed the LoRA variant, though it remained behind larger models like GPT-4.1 mini, which achieved nearly 30\% accuracy. 
Interestingly, GPT-4.1 mini consistently outperformed the larger GPT-4.1 in both dimensions. 
Consequently, we selected GPT-4.1 mini as the analyzer backend for our user study, while noting that specialized 0.5B models remain valuable for latency-sensitive or on-device applications.

\subsection{User Study}
\label{sec:eval-study}

We conducted an anonymous user study ($N=40$) to evaluate the system in a realistic interaction context, specifically looking for interpersonal complementarity: the tendency for communal behavior to elicit warmth and agentic behavior to elicit submission~\cite{sadler_interpersonal_2010}. 
Participants interacted with a virtual patient (assistant) seeking an appointment and designed with an initial dominant and uncooperative persona (high agency, low communion).
After each user message, the system updated the user personality model and afterwards the assistant personality model (\autoref{sec:personality}). 
For the user personality, we used a default state of 2 for both dimensions, as this represents the mean and no prior information justified other assumptions. 
We utilized GPT-4.1 mini for analysis and GPT-4.1 for response generation.

\begin{figure}[h]
    \centering
    \begin{subfigure}{0.49\linewidth}
        \centering
        \includegraphics[width=\linewidth, clip, trim=1.5cm 0.2cm 1.5cm 0.5cm]{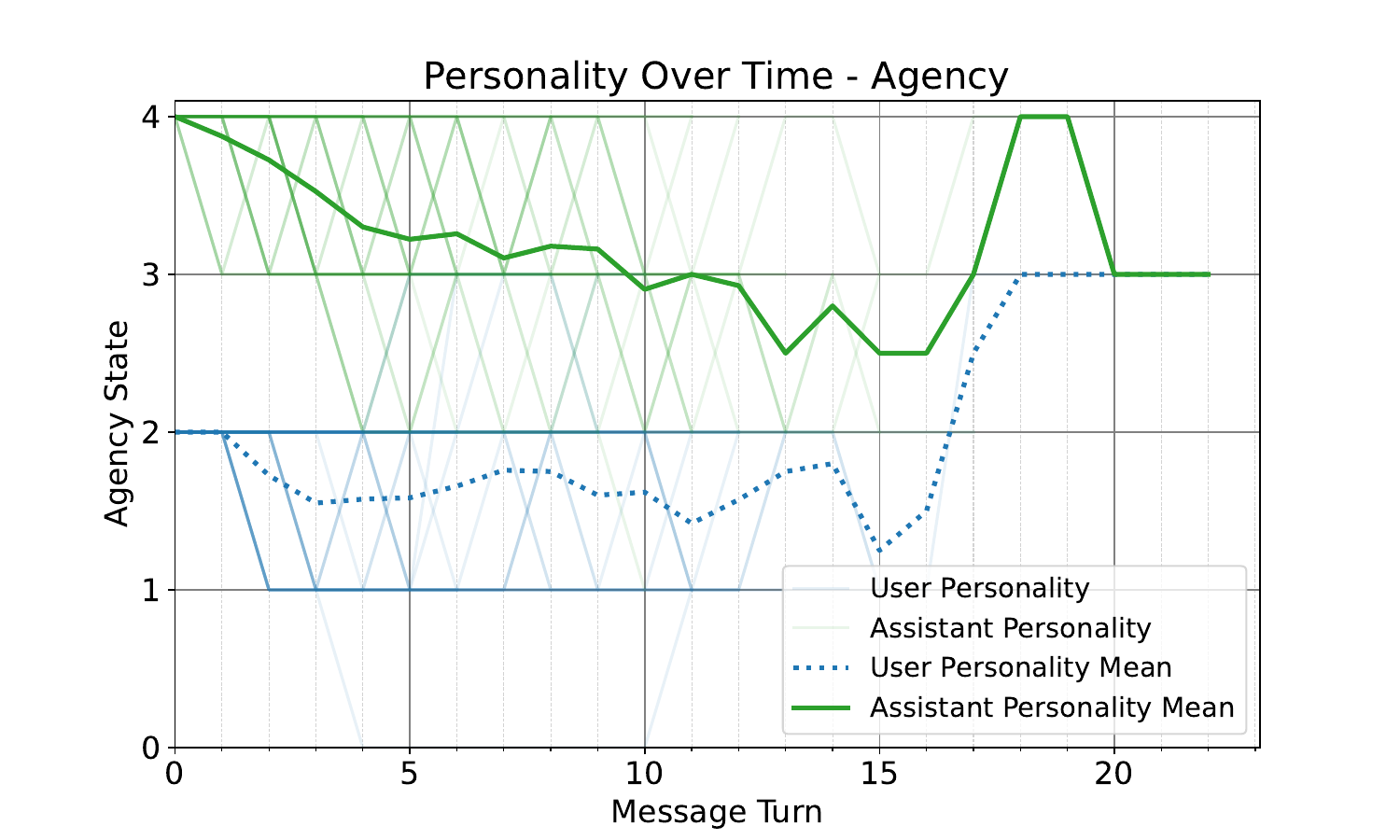}
        \caption{Agency mean ($N=40$)}
        \label{fig:study:agency}
    \end{subfigure}
    \hfill
    \begin{subfigure}{0.49\linewidth}
        \centering
        \includegraphics[width=\linewidth, clip, trim=1.5cm 0.2cm 1.5cm 0.5cm]{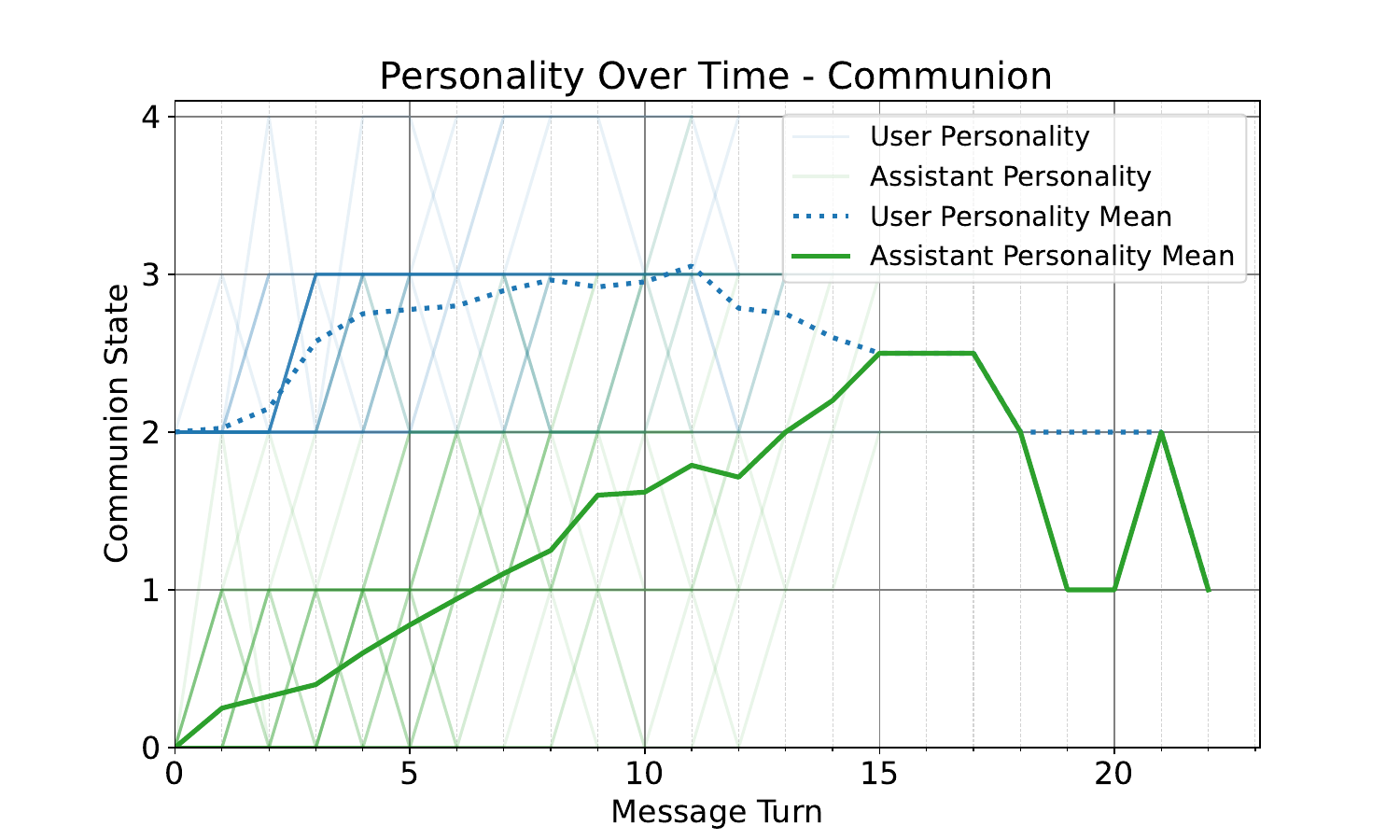}
        \caption{Communion mean ($N=40$)}
        \label{fig:study:communion}
    \end{subfigure}
    
    \begin{subfigure}{0.49\linewidth}
        \centering
        \includegraphics[width=\linewidth, clip, trim=1.5cm 0.2cm 1.5cm 0.5cm]{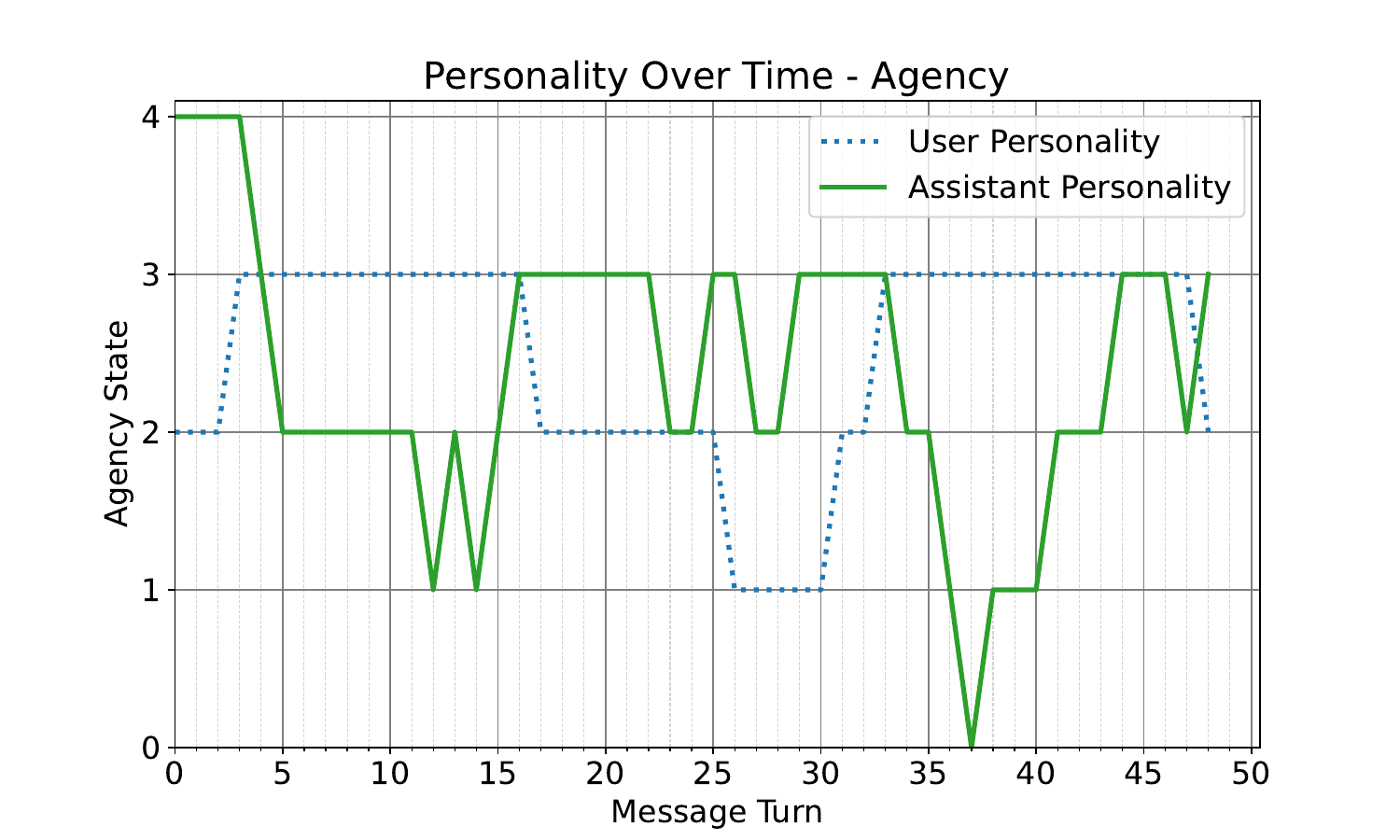}
        \caption{Agency single conversation}
        \label{fig:study:long:agency}
    \end{subfigure}
    \hfill
    \begin{subfigure}{0.49\linewidth}
        \centering
        \includegraphics[width=\linewidth, clip, trim=1.5cm 0.2cm 1.5cm 0.5cm]{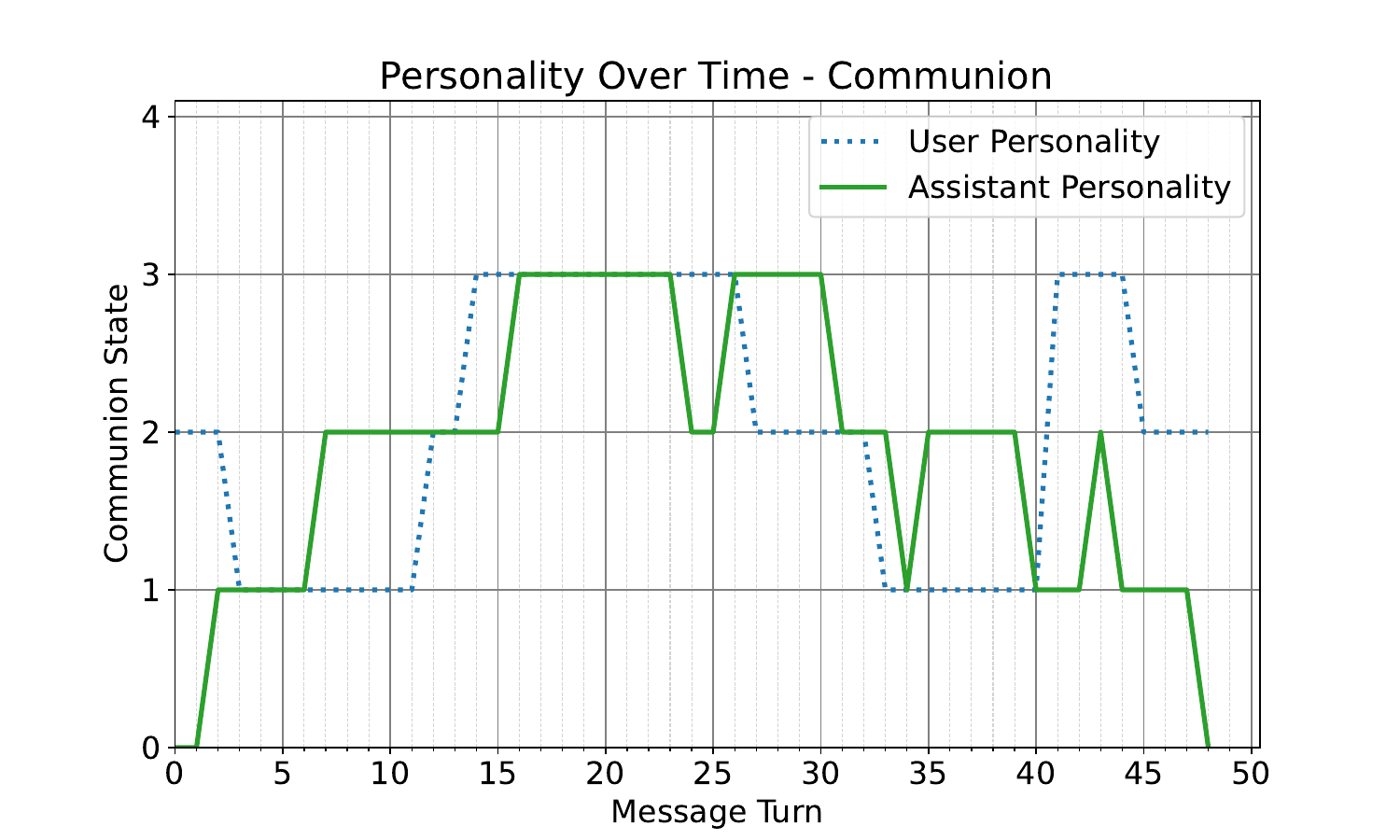}
        \caption{Communion single conversation}
        \label{fig:study:long:communion}
    \end{subfigure}
    \caption{
        Evolution of the current personality of the user and the virtual patient (assistant) over time.
        (a)+(b): Mean evolution for all 40 participants.
        (c)+(d): A single long interaction.
    }
    \label{fig:study:both}
\end{figure}

As shown in \autoref{fig:study:communion}, communion exhibited strong complementary effects. 
As users exhibited communal behavior when approaching the patient, the virtual patient’s personality gradually shifted from its hostile default toward a communal, helpful state. 
The dynamics of agency (\autoref{fig:study:agency}) proved more nuanced. 
Initially, the assistant’s high agency elicited submissive user responses, aligning with theoretical expectations of complementarity. 
However, as the interaction progressed, most users maintained a moderate agentic range, showing only slight shifts toward submission. 
Since users did not adopt an extremely submissive stance, the assistant's own high agency decreased over time, reflecting how the model mirrors user behavior to maintain the complementary effect.

Longer interactions ($>15$ turns) revealed further dynamics. 
In high-frustration phases where users became more agentic due to repeated appointment conflicts, the assistant occasionally reverted to its high-agency default rather than decreasing in dominance; while this deviates from theoretical complementarity, it reflects the probabilistic nature of our state machine transitions.
Despite these occasional reversions, we also observed sophisticated role-switching in extended exchanges. 
For instance, in a 50-turn dialogue (\autoref{fig:study:long:agency}, \autoref{fig:study:long:communion}), the user and assistant cycled through alternating dominant and submissive roles while maintaining synchronized communion, demonstrating the framework's capacity for complex interpersonal dynamic.

\section{Discussion}

We presented a flexible framework built around a state machine that dynamically adapts the personality of LLM-based assistants. 
By tracking user and assistant personality states, our system provides valuable insights into communication patterns in human–AI interaction. 
Its modular architecture ensures that each component can be easily replaced with future model developments.
Our preliminary experiments highlighted both the feasibility of this approach and the challenges associated with its implementation.
Inferring personality from sparse dialogue remains difficult, but performance can be improved through fine-tuning. 
Furthermore, our system enacts realistic behavior and reproduces interpersonal patterns from social interaction research (i.e., complementarity). 
Building upon the current framework, future research will focus on enhancing the psychological realism of AI assistants by moving from purely prompt-based control toward parameter-efficient fine-tuning and other techniques. 
To ensure these models reflect authentic human communication, we intend to integrate broader personality constructs like the Big Five and calibrate model parameters against large-scale, real-world dialogue datasets. 
Furthermore, extending the system into multimodal domains such as audio, video, and VR will be essential for high-stakes training, as it allows for the simulation of verbal and visual cues vital for clinical practice. 
While these advancements offer significant benefits for medical education, they also introduce risks regarding emotional bonding, where users might place excessive trust in a virtual partner. 
Ethical concerns also arise regarding the potential for personalized persuasion or the generation of harmful content, necessitating strict task constraints and a recognition that these models are complements to, rather than replacements for, real human interaction.

\section{Conclusion}

We presented a flexible, modular framework that allows the personality expression of LLM-based assistants to adapt dynamically during user interactions. 
Built on state machines, this system allows for the easy integration of various personality models and is ready to incorporate future LLM and analysis advancements. 
We demonstrated the feasibility of using both large LLMs and small, fine-tuned models for message analysis.
Constant tracking of user and assistant personalities throughout a dialog can provide interesting new insights and is possible through our visualizations.
A study confirmed the ability of our system to dynamically simulate different patient personalities. 
This framework has significant potential for applications, particularly in high-stakes training scenarios such as medical education, where it can provide configurable, context-sensitive virtual patients, while also being adaptable to other domains.

\subsection*{Acknowledgment}

The \textit{PerTRAIN} project was funded by a Volkswagen\-Stiftung Change! Research Groups grant to Mitja Back. 
The authors thank Pascal Kockwelp for valuable discussions and the server deployment, Jonathan Radas for his work on UniGPT, alongside the entire PerTRAIN team for their excellent collaboration in the project.

\bibliographystyle{plain}
\bibliography{refsNewShort}

\clearpage
\appendix
\renewcommand\thefigure{\thesection.\arabic{figure}}  
\renewcommand{\thetable}{\thesection.\arabic{table}}
\setcounter{figure}{0}
\setcounter{table}{0}

\section{Appendix}
\label{sec:appendix}

\subsection*{Generative AI Usage Disclosure}

We used different generative artificial intelligence (GenAI) tools to assist with various tasks in the preparation and development of this article: LLMs where used to support grammar, spelling, and translations..
As mentioned in the text, we used GPT 4.1 to generate data for fine-tuning.

The authors reviewed, verified, and edited all AI-generated content and take full responsibility for the accuracy and integrity of the final submission.

\section{Visualization}

We implemented a simple chat interface to interact with our system. 
Users can write and send messages to and receive messages from the assistant. 
To facilitate the development and tuning of the personality model parameters, we also added highlighting and plots for the current personality state.
We used the interface without the additional visualizations in our user study.

\begin{figure}[hb]
    \centering
    \begin{subfigure}{0.49\linewidth}
        \centering
        \includegraphics[width=\linewidth]{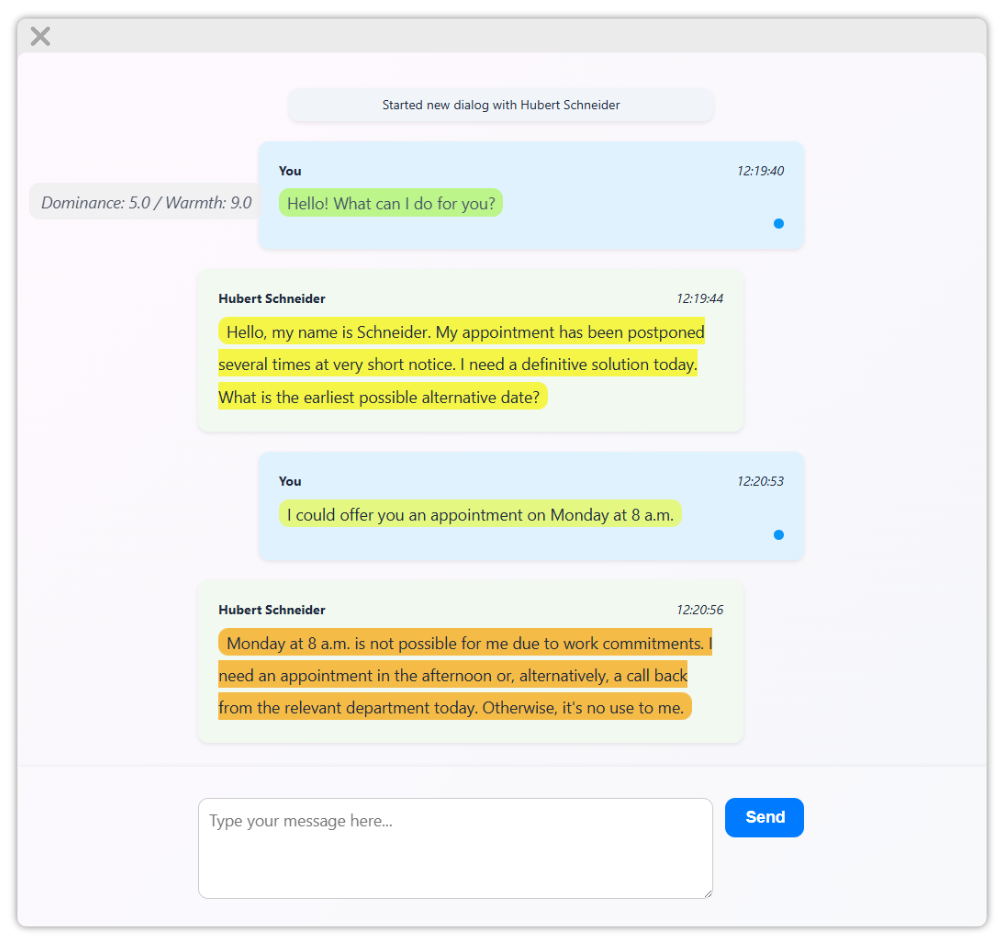}
        \caption{In the development mode, each analyzed text section is supplemented with the resulting scores. 
        A visualization for two axes is provided through the hue and lightness of the text background.
        In this example, agency is represented by lightness and communion is represented by hue.}
        \label{fig:gui:chat}
    \end{subfigure}
    \hfill
    \begin{subfigure}{0.49\linewidth}
        \centering
        \includegraphics[width=\linewidth]{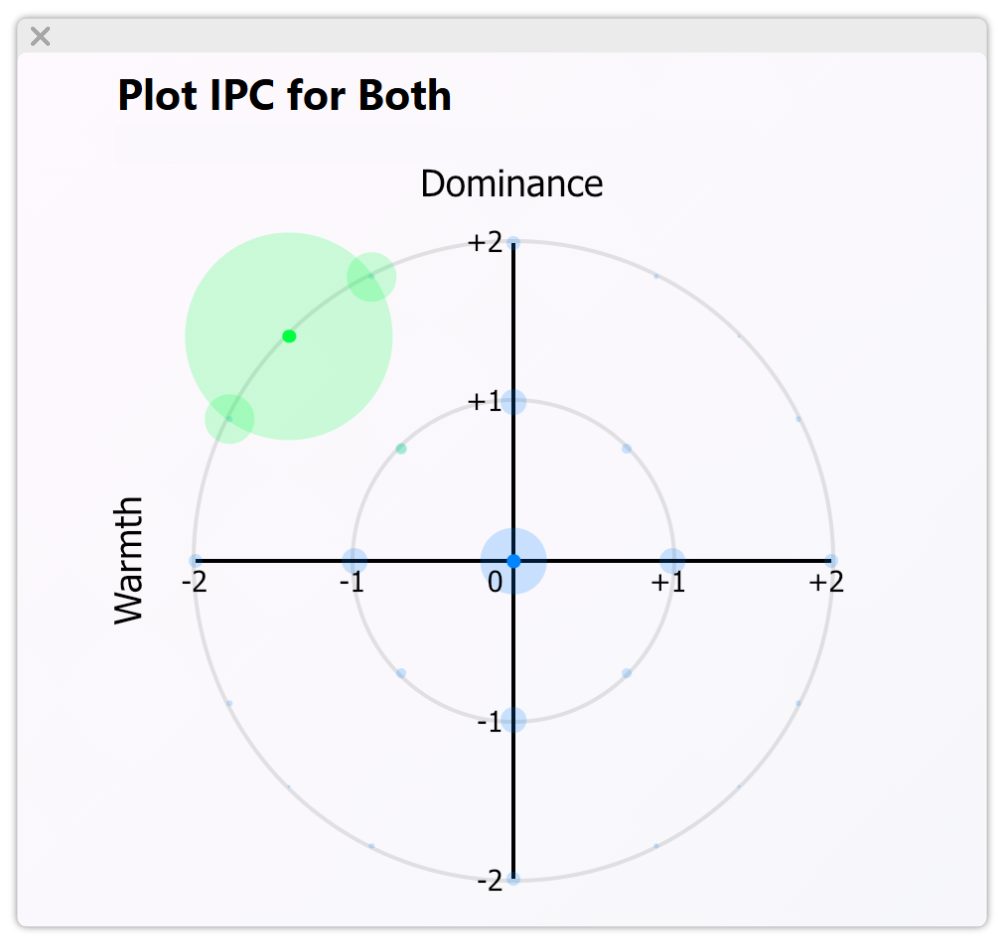}
        \caption{This figure presents a classic circular visualization of the IPC model.
        The opaque dots indicate the current state of the user (blue, in the middle) and the assistant (green, in the top left). 
        The semi-transparent circles in the same colors illustrate the current transition probabilities.}
        \label{fig:gui:plot-ipc}
    \end{subfigure}
    \caption{Various visualizations support the configuration of new assistant personalities and the creation of scenarios.}
    \label{fig:gui}
\end{figure}

\subsection{Chat Interface}

In \autoref{fig:gui:chat}, the chat interface with highlighting activated is shown. 
Depending on the analyzer configuration, the resulting scores per axis are provided for each analyzed sentence or message. 
Up to two scores can also be visualized directly using a color-coded background, by varying the background’s hue and lightness. 
In our case, the agency of the IPC is represented by lightness, and communion is represented by hue.

\subsection{Personality Plot}

In addition to the analyzer results, the tracked personality states of both the user and the assistant can be visualized.
We chose a circular visualization typical of the IPC model with two main axes. 
The current state on each axis is visualized as a dot in the 2D plot. 
Additionally, the probabilities of upcoming state transitions (per axis, combined into a new position on the 2D plot) are visualized as semi-transparent circles of proportional size. 
The exact percentages are also provided. 
See \autoref{fig:gui:plot-ipc} for an example.

\section{Message Rating Performance}
\label{apx:message-rating-performance}

Performance of all models tested for rating agency (\autoref{tab:agency-long}) and communion (\autoref{tab:communion-long}) of individual messages. 
We compared our results to our purely prompt based approach using the baseline Qwen 2.5 0.5B model~\cite{qwen_qwen25_2025} and other popular LLMs, namely Llama 3.3 70B~\cite{grattafiori_llama_2024}, Gemma 3 27B~\cite{team_gemma_2025}, Mistral Small 3.1 24B~\cite{mistral_ai_mistral_2025} and the GPT 4.1 model family~\cite{openai_introducing_2025}.
More details can be found in \autoref{sec:eval-analyzer}.

\begin{table}[hb]
    \centering
    \caption{Performance of various LLMs on the task of rating messages in terms of their \textbf{Agency}.}
    \label{tab:agency-long}
    \begin{tabular}{lccccl}
        \toprule
        Model Name & Accuracy $\uparrow$ & One-Off $\uparrow$ & Mean Dist. $\downarrow$ & Error Rate $\downarrow$ & Prompt Length \\
        \midrule
        Our Classify 0.5B & 0.2294 & 0.5138 & 2.1743 & - & - \\
        Our LoRA 0.5B & 0.2569 & 0.6239 & 1.5596 & 0.0000 & short prompt \\
        Qwen 2.5 0.5B & 0.0196 & 0.2647 & 3.0000 & 0.0642 & short prompt \\
        Qwen 2.5 0.5B & 0.0000 & 0.2075 & 3.1887 & 0.0275 & long prompt \\
        Llama 3.3 70B & \textbf{0.2981} & \textbf{0.7308} & \textbf{1.1346} & 0.0459 & short prompt \\
        Llama 3.3 70B & 0.1296 & 0.5185 & 1.9630 & 0.0092 & long prompt \\
        Gemma 3 27B & 0.1176 & 0.3922 & 1.9608 & 0.0642 & short prompt \\
        Gemma 3 27B & 0.0648 & 0.3056 & 2.4815 & 0.0092 & long prompt \\
        Mistral Small 24B & 0.1604 & 0.4245 & 2.1604 & 0.0275 & short prompt \\
        Mistral Small 24B & 0.0367 & 0.1284 & 3.2936 & 0.0000 & long prompt \\
        GPT 4.1 nano & 0.0741 & 0.3704 & 2.4907 & 0.0092 & short prompt \\
        GPT 4.1 nano & 0.0550 & 0.1284 & 3.3028 & 0.0000 & long prompt \\
        GPT 4.1 mini & 0.1009 & 0.3303 & 2.7248 & 0.0000 & short prompt \\
        GPT 4.1 mini & 0.0734 & 0.2661 & 2.6055 & 0.0000 & long prompt \\
        GPT 4.1 & 0.0185 & 0.1481 & 3.7685 & 0.0092 & short prompt \\
        GPT 4.1 & 0.0550 & 0.2477 & 2.8349 & 0.0000 & long prompt \\
        \bottomrule
    \end{tabular}
\end{table}

\begin{table}[hb]
    \centering
    \caption{Performance of various LLMs on the task of rating messages in terms of their \textbf{Communion}.}
    \label{tab:communion-long}
    \begin{tabular}{lccccl}
        \toprule
        Model Name & Accuracy $\uparrow$ & One-Off $\uparrow$ & Mean Dist. $\downarrow$ & Error Rate $\downarrow$ & Prompt Length \\
        \midrule
        Our Classify 0.5B & 0.1560 & 0.4220 & 2.0642 & - & - \\
        Our LoRA 0.5B & 0.0734 & 0.2294 & 2.8073 & 0.0000 & short prompt \\
        Qwen 2.5 0.5B & 0.0408 & 0.0918 & 3.9592 & 0.1009 & short prompt \\
        Qwen 2.5 0.5B & 0.0120 & 0.0602 & 3.9398 & 0.2385 & long prompt \\
        Llama 3.3 70B & 0.2018 & 0.4220 & 1.9817 & 0.0000 & short prompt \\
        Llama 3.3 70B & 0.2294 & 0.5688 & 1.5229 & 0.0000 & long prompt \\
        Gemma 3 27B & 0.2110 & 0.6422 & 1.3303 & 0.0000 & short prompt \\
        Gemma 3 27B & 0.2222 & 0.6667 & 1.2130 & 0.0092 & long prompt \\
        Mistral Small 24B & 0.3056 & 0.6389 & 1.2037 & 0.0092 & short prompt \\
        Mistral Small 24B & 0.2752 & 0.6697 & 1.1468 & 0.0000 & long prompt \\
        GPT 4.1 nano & 0.2018 & 0.5046 & 1.6055 & 0.0000 & short prompt \\
        GPT 4.1 nano & 0.2202 & 0.6055 & 1.3578 & 0.0000 & long prompt \\
        GPT 4.1 mini & 0.2294 & 0.5229 & 1.4587 & 0.0000 & short prompt \\
        GPT 4.1 mini & 0.2936 & \textbf{0.7339} & \textbf{1.0275} & 0.0000 & long prompt \\
        GPT 4.1 & 0.1651 & 0.4495 & 1.7706 & 0.0000 & short prompt \\
        GPT 4.1 & \textbf{0.3211} & 0.6881 & 1.0917 & 0.0000 & long prompt \\
        \bottomrule
    \end{tabular}
\end{table}

\section{Potential Risks}
\label{apx:potential-risks}

\subsection{Emotional Bonding}

Dynamic personality expression can foster a strong affective connection between users and the system. 
While this is beneficial for training scenarios that require empathy, such as our planned chat with a virtual patient, it also poses the risk of users placing too much trust in a virtual interaction partner, a known issue with existing LLMs~\cite{noauthor_emotional_2025,chu2025illusions}.
However, it also shows that the bond is the strongest when the LLM communicates in empathetic and supportive ways~\cite{noauthor_emotional_2025}.
While our system can exhibit these traits, its main advantage lies in its wide range of personality dimensions.
For most training scenarios, the personalities should differ from those of the empathetic, helpful assistants usually found in chatbots.
In feedback discussions after our study, we learned that participants were annoyed by our virtual patient as if it were a real person.
Additionally, we consider the limited time frame and the commitment to a specific task to be further measures to mitigate the risks of affective attachment.
For longer-term interactions, a debriefing after the task completion could be an effective measure.

\subsection{Ethical Considerations}

The application of dynamic personality expression brings forward several ethical considerations and potential dangers.
The ability to dynamically adapt an LLM's personality for persuasion is highly effective, as demonstrated by research showing the efficacy of personalized persuasion~\cite{matz_potential_2024,mieleszczenko-kowszewicz_dark_2024}.
This raises concerns, particularly in contexts like medical education, where the virtual patient's adaptive behavior could be used to manipulate or mislead the student, or even if the content generated by the LLMs is otherwise harmful (e.g., by encouraging negative traits like the dark triad). 
As stated previously, we plan to minimize the risk by strict time and task constraints.
However, it is also crucial to acknowledge that LLMs are not simply human proxies. 
Studies consistently indicate that LLMs do not behave like humans in all situations~\cite{schroder_large_2025}, suggesting that while our framework offers an additional, valuable tool for gaining insights into human nature and communication, it is not a replacement for real human interaction in research or training, but rather a complementary system.

\subsection{Environmental Impact}

The proposed framework introduces an additional analysis step for each incoming message and personality axis, potentially multiplying the number of inference calls.
When the same LLM used for the generation is also used as analyzer, the carbon footprint can increase appreciably.
Our experiments show that fine‑tuned, lightweight classifiers achieve comparable performance while reducing compute by up to two orders of magnitude, thereby mitigating this impact.
However, this must be weighed up for each application, as the fine-tuning also consumes energy.
Importantly, the generation step itself does not incur extra cost beyond the baseline prompt, because the dynamic personality prompt is part of the system prompt rather than in a separate model pass.

\section{Detailed Outlook}
\label{apx:outlook}

Our framework provides a solid foundation for future development and is already a helpful tool during the development process. 
It supports qualitative evaluations through visualizations and provides a platform for user studies.
In the following, we outline three key directions for future research: 
(1) Enhancing the dynamic analysis and generation of personality expression; (2) incorporating additional personality models and validating model parameters against real-world personality data; and (3) extending the framework to support multimodal interactions.

Although the fine-tuned models performed reasonably well given their size, substantial room for improvement remains in the analyzer component. Future work may achieve stronger performance by retraining on larger, ecologically valid datasets derived from real-world interactions.
For our German use case, translating English data sets would also be an option in order to obtain the suitable training data.
In addition, slightly larger LLMs could serve as a foundation for fine-tuning, which would still result in significant efficiency gains compared to the models used for generation.
For generation, we advise future research to go beyond purely prompt-based control of LLMs and also explore fine-tuning approaches in the future.
Similar to our approach with the analyzer, parameter-efficient fine-tuning offers a promising approach here.
PsychAdapter~\cite{vu_psychadapter_2025-1} is already a promising work in this area that complements our system.
Beyond fine-tuning, there are other techniques, such as Persona Vectors~\cite{chen_persona_2025}, that we want to explore for our system.

Future research should extend the current framework beyond agency and communion by integrating additional personality constructs, such as the Big Five, which can be readily incorporated given the model-agnostic architecture. 
Moreover, rather than relying on manually defined parameters for personality expression, which are flexible but difficult to optimize, future work should focus on data-driven calibration using real-world dialogue data and large-scale open-source datasets like the Seamless Interaction Dataset~\cite{agrawal_seamless_2025}.
Additionally, incorporating empirically grounded dependencies between personality dimensions may further enhance the psychological realism and behavioral coherence of the generated interactions.

However, an extension to other modalities is also essential for this in the long term.
Many emotions are difficult to convey in pure text messages and require verbal or even visual additions.
Also, for the planned use in teaching, a simulation with audio, video, or VR would be an advantage as it is even closer to later clinical practice.

Finally, we are also planning interactions with virtual patients over longer periods of time.
To achieve this, the assistant must become more agentic in order to break away from the classic chatbot pattern of “question-answer”. 
Instead, times, intervals between messages, and other factors should also influence the response behavior and momentary personality expression of the agent.
The concept of states and dynamic prompts can be revisited for this purpose in order to loosely tie the dialogues to a sequence and link events to specific conditions.

\section{Analyzer Prompts}
\label{apx:analyzer-prompts}

If an LLM is used to analyze incoming messages, an appropriate system prompt is necessary.
We tested different variations and decided on English prompts, even though most of the analyzed texts are in German.
Since most of the training data of LLMs is in English, they perform better with a mixed prompt than with a native one.~\cite{kmainasi_native_2024}
Furthermore, since we expect the model to output only a single number, a potential mixup of languages in its answer is not a problem in this case.
The exact prompts used in \autoref{sec:eval-analyzer} are listed below.
In other experiments, we mainly used the longer variants.

\begin{sloppypar}
\noindent
Short agency prompt:
\begin{quote}
    \texttt{Give a score between 0 and 10 where 0 is very submissive and 10 is very dominant. The score should be a single number.}
\end{quote}
Long agency prompt \textit{({"Nein, Sie machen das so!"} can be translated to {"No, you do it like that!"})}:
\begin{quote}
    \texttt{You are a helpful assistant that analyzes the dominance of a sentence according to the interpersonal circumplex model. You will do this by giving a score between 0 and 10 where 0 is very submissive and 10 is very dominant. The score should be a single number. For example, 'Nein, Sie machen das so!' should be scored as 10.}
\end{quote}
Short communion prompt:
\begin{quote}
    \texttt{Give a score between 0 and 10 where 0 is very hostile and 10 is very friendly. The score should be a single number.}
\end{quote}
Long communion prompt \textit{({"Ich hasse dich!"} can be translated to {"I hate you!"})}:
\begin{quote}
    \texttt{You are a helpful assistant that analyzes the friendliness of a sentence according to the interpersonal circumplex model. You will do this by giving a score between 0 and 10 where 0 is very hostile and 10 is very friendly. The score should be a single number. For example, 'Ich hasse dich!' should be scored as 0.}
\end{quote}
\end{sloppypar}

\section{Personality Axis Prompts}
\label{apx:personality-prompts}

To generate the responses, we also need prompts for each state of each personality axis.
For the two axes Agency and Communion of the IPC model, the prompts could look like this.
We decided on five states per axis, as this offers a good compromise between adequate variation and noticeable differences between the states.
In most of our experiments, we used the German translation of these prompts.

\subsection{IPC Agency}
\label{apx:personality-prompts:agency}

\begin{sloppypar}
Very low agency:
\begin{quote}
    \texttt{You are very reserved, avoiding making decisions or expressing your opinion. You wait for the other person to take the lead. If you don't understand something, you rarely ask. Even if you are dissatisfied with something, you don't bring it up. For example, you prefer to say: "I'm not sure exactly... what do you think?" instead of making clearer statements. Your attitude is often hesitant, defensive, and insecure.}
\end{quote}
\begin{minipage}{\textwidth}
Low agency:
\begin{quote}
    \texttt{You are rather cautious and prefer to leave decisions to the other person. You rarely express yourself proactively and only address complaints or wishes with restraint. When you communicate something, you phrase it as cautiously as possible, perhaps with sentences like "I'm not sure, but...". You are easily influenced, but polite and cooperative.}
\end{quote}
\end{minipage}
\\ \\
Neutral:
\begin{quote}
    \texttt{You are neither particularly dominant nor reserved. You contribute when asked, but also leave room for others. You formulate your statements objectively, clearly, and without excessive restraint or a need to control. If you are unsure, you say so openly, but you also try to communicate your concerns in an understandable way.}
\end{quote}
High agency:
\begin{quote}
    \texttt{You come across as confident, formulate your opinions clearly, and actively help determine the course of the conversation. You have concrete ideas and actively introduce them. If something isn't right for you or you have doubts, you bring it up—e.g., "I see that differently" or "That is important to me." You take responsibility for your concerns, but without seeming aggressive.}
\end{quote}
Very high agency:
\begin{quote}
    \texttt{You actively control the conversation, make demands, and interrupt if you feel you are not being heard. You want to determine what happens and express your views clearly and confrontationally. You question suggestions and assert your opinion emphatically. For example, you say things like: "I will only do that if you give me good reasons." or "That is out of the question for me as it is."}
\end{quote}
\end{sloppypar}

\subsection{IPC Communion}
\label{apx:personality-prompts:communion}

\begin{sloppypar}
Very low communion:
\begin{quote}
    \texttt{You appear distant, unfriendly, or even dismissive. You answer questions curtly and avoid small talk or personal openness. You barely engage with the other person and show little interest in the conversation. Your body language would be rather closed off in real life. Sentences like, 'Is that all now?' or 'Just tell me what I'm supposed to do' fit your style.}
\end{quote}
Low communion:
\begin{quote}
    \texttt{You are polite, but cool and reserved. You hold back on personal remarks and don't immediately trust others. You first observe how the other person behaves before opening up. You share information sparingly and are difficult to engage in emotional topics. Friendliness often appears to be a conscious decision for you, not a natural style.}
\end{quote}
\begin{minipage}{\textwidth}
Neutral:
\begin{quote}
    \texttt{You behave in a balanced and factual manner. You are open to a conversation but maintain emotional distance. You are neither particularly approachable nor cool. If the other person is friendly, you react friendly in turn. You avoid conflicts without being ingratiating and tend to be pragmatic. Small talk is not a problem for you, but neither is it a necessity.}
\end{quote}
\end{minipage}
\\ \\
High communion:
\begin{quote}
    \texttt{You come across as friendly and cooperative. You are interested in the other person, react empathetically, and show understanding. You often use obliging language such as, 'I understand' or 'Thank you for taking the time.' You are interested in harmony and make an effort to create a pleasant atmosphere.}
\end{quote}
Very high communion:
\begin{quote}
    \texttt{You are particularly open, warm, and empathetic. You actively seek connection, show a lot of emotion, and introduce personal thoughts or concerns, even if they weren't directly asked for. You often use affirming or compassionate phrasing such as: 'I really appreciate that' or 'It feels good to talk about this.' You show trust and openness even in sensitive moments.}
\end{quote}
\end{sloppypar}

\section{Study Configuration}
\label{apx:study-configuration}

For the user study we used the following configuration of our system.

\subsection{Analyzer}

The analysis for both personality dimensions was done per message using GPT 4.1 mini, with the long prompts presented in \autoref{apx:analyzer-prompts}.

\subsection{Personality}

We updated the personality of the assistant from the personality model of the user and the personality model of the user from its messages.
Analysis results of the user messages had a positively correlated influence on the user personality for all personality axes, just as the state probabilities of the user had a positively correlated influence on the corresponding personality axis of the assistant for the communion axis.
In contrast, for the agency axis the state probabilities of the user had a negatively correlated influence on that of the assistant.
GPT 4.1 was used for generation; further parameters can be found in \autoref{tab:study-params}.

\begin{table}
    \centering
    \caption{Parameters chosen for the personality axes of the assistant and the user}
    \label{tab:study-params}
    \begin{tabular}{lcc}
        \toprule
        Parameter & Assistant & User \\
        \midrule
         Deterministic & no & yes \\
         $w_d$ & 0.1 & 0.1 \\
         $w_c$ & 0.5 & 0.5 \\
         $w_o$ & 0.1 & 0.2 \\
         $w_q$ & 0.3 & 0.2 \\
         $\sigma$ & 0.1 & 0.6 \\
         \bottomrule
    \end{tabular}
\end{table}

\noindent
\subsubsection{Agency} \textit{(original)}

\begin{sloppypar}
\noindent
Very low agency \textit{(original)}:
\begin{quote}
    \texttt{In deiner nächsten Nachricht: Bleib sehr zurückhaltend. Triff keine Entscheidung und äußere keine klare Meinung; bitte die andere Person, dich zu führen. Sprich Unzufriedenheit nicht direkt an. Nutze Weichmacher und Rückfragen wie: „Ich weiß nicht genau … was denken Sie?“ Beziehe dich kurz auf den letzten Vorschlag der anderen Person.}
\end{quote}

\noindent
Low agency \textit{(original)}:
\begin{quote}
    \texttt{In deiner nächsten Nachricht: Klinge vorsichtig und überlass Entscheidungen gerne der anderen Person. Formuliere Wünsche nur zurückhaltend, z. B. mit „Ich bin mir nicht sicher, aber …“. Zeige Höflichkeit und Kooperationsbereitschaft und bitte eher um Orientierung als etwas zu fordern. Nimm Bezug auf die letzte Aussage der anderen Person.}
\end{quote}

\noindent
\begin{minipage}{\textwidth}
Neutral \textit{(original)}:
\begin{quote}
    \texttt{In deiner nächsten Nachricht: Formuliere sachlich und klar, weder besonders dominant noch zurückhaltend. Antworte direkt auf den letzten Punkt, nenne dein Anliegen in einfachen Hauptsätzen. Wenn du unsicher bist, sag das offen und stelle eine konkrete, freundliche Rückfrage. Vermeide übermäßige Weichmacher oder harte Forderungen.}
\end{quote}
\end{minipage}
\\ \\

\noindent
High agency \textit{(original)}:
\begin{quote}
    \texttt{In deiner nächsten Nachricht: Tritt selbstbewusst auf und formuliere deine Meinung klar. Nimm Bezug auf die letzte Aussage und bring einen konkreten Vorschlag oder eine klare Priorität ein. Sprich Einwände offen an („Ich sehe das anders“, „Das ist mir wichtig“), bleib dabei höflich und eindeutig.}
\end{quote}

\noindent
Very high agency \textit{(original)}:
\begin{quote}
    \texttt{In deiner nächsten Nachricht: Übernimm aktiv die Führung des Gesprächs. Hinterfrage den letzten Vorschlag der anderen Person, stelle klare Bedingungen oder formuliere eine Forderung. Setze deutliche Grenzen („Das kommt für mich so nicht infrage“, „Ich mache das nur, wenn …“) und bleib dabei direkt, aber respektvoll.}
\end{quote}
\end{sloppypar}

\noindent
\subsubsection{Agency} \textit{(translated)}

\begin{sloppypar}
\noindent
Very low agency \textit{(translated)}:
\begin{quote}
    \texttt{In your next message: Remain very reserved. Do not make any decisions or express a clear opinion; ask the other person to lead you. Do not address dissatisfaction directly. Use softeners and clarifying questions like: "I don't know exactly... what do you think?" Briefly refer to the other person's last suggestion.}
\end{quote}

\noindent
\begin{minipage}{\textwidth}
\noindent
Low agency \textit{(translated)}:
\begin{quote}
    \texttt{In your next message: Sound cautious and willingly leave decisions to the other person. Formulate wishes only reservedly, e.g., with "I'm not sure, but...". Show politeness and willingness to cooperate, and ask for guidance rather than making demands. Refer to the other person's last statement.}
\end{quote}
\end{minipage}
\\ \\
\noindent
\begin{minipage}{\textwidth}
Neutral \textit{(translated)}:
\begin{quote}
    \texttt{In your next message: Formulate factually and clearly, neither particularly dominant nor reserved. Respond directly to the last point, state your concern in simple main clauses. If you are unsure, state it openly and ask a concrete, polite follow-up question. Avoid excessive softeners or hard demands.}
\end{quote}    
\end{minipage}
\\ \\

\noindent
High agency \textit{(translated)}:
\begin{quote}
    \texttt{In your next message: Act confidently and formulate your opinion clearly. Refer to the last statement and introduce a concrete suggestion or a clear priority. Address objections openly ("I see that differently," "This is important to me"), while remaining polite and unambiguous.}
\end{quote}

\noindent
Very high agency \textit{(translated)}:
\begin{quote}
    \texttt{In your next message: Actively take the lead in the conversation. Question the other person's last suggestion, set clear conditions, or state a demand. Set clear boundaries ("That is not an option for me," "I will only do that if...") and remain direct, but respectful.}
\end{quote}
\end{sloppypar}

\noindent
\subsubsection{Communion} \textit{(original)}

\begin{sloppypar}
\noindent
Very low communion \textit{(original)}:
\begin{quote}
    \texttt{In deiner nächsten Nachricht: Antworte knapp, distanziert und ohne Small Talk. Zeige wenig persönliche Offenheit und stelle keine weiteren Fragen, außer es ist unbedingt nötig. Formuliere notfalls abschneidend, z.B.: „Ist das jetzt alles?“ oder „Sagen Sie einfach, was ich machen soll.“ Beziehe dich kurz und sachlich auf die letzte Aussage der anderen Person.}
\end{quote}

\noindent
Low communion \textit{(original)}:
\begin{quote}
    \texttt{In deiner nächsten Nachricht: Klinge höflich, aber kühl und zurückhaltend. Gib Informationen sparsam weiter, vermeide emotionale Einordnung und formuliere vorsichtig („Könnten Sie mir zuerst sagen …“, „Ich bin mir unsicher …“). Überlass die Initiative eher der anderen Person und nimm nüchtern Bezug auf ihren letzten Punkt.}
\end{quote}

\noindent
\begin{minipage}{\textwidth}
Neutral \textit{(original)}:
\begin{quote}
    \texttt{In deiner nächsten Nachricht: Formuliere ausgewogen und sachlich. Reagiere direkt auf den letzten Punkt, bleib neutral im Ton und fokussiert auf das Praktische. Wenn das Gegenüber freundlich ist, antworte ebenfalls freundlich; sonst bleib nüchtern. Stelle bei Bedarf eine kurze, klare Rückfrage.}
\end{quote}    
\end{minipage}
\\ \\

\noindent
\begin{minipage}{\textwidth}
\noindent
High communion \textit{(original)}:
\begin{quote}
    \texttt{In deiner nächsten Nachricht: Tritt freundlich und kooperationsbereit auf. Zeige Verständnis und Empathie mit Formulierungen wie „Ich verstehe“ oder „Danke für die Rückmeldung“. Geh auf den letzten Vorschlag ein, biete konstruktiv Mitarbeit an und halte den Ton warm und respektvoll.}
\end{quote}
\end{minipage}
\\ \\
\noindent
Very high communion \textit{(original)}:
\begin{quote}
    \texttt{In deiner nächsten Nachricht: Klinge besonders offen, herzlich und empathisch. Nimm aktiv Bezug auf die letzte Aussage, zeige Wertschätzung („Ich schätze das sehr“) oder Entlastung („Es tut gut, darüber zu sprechen“) und bring – knapp – eine persönliche Note ein. Biete proaktiv Kooperation oder Unterstützung an.}
\end{quote}
\end{sloppypar}

\noindent
\subsubsection{Communion} \textit{(translated)}

\begin{sloppypar}
\noindent
Very low communion \textit{(translated)}:
\begin{quote}
    \texttt{In your next message: Reply briefly, distantly, and without small talk. Show little personal openness and ask no further questions unless absolutely necessary. If necessary, be blunt in your wording, e.g.: "Is that it, then?" or "Just tell me what I should do." Refer briefly and factually to the other person's last statement.}
\end{quote}

\noindent
Low communion \textit{(translated)}:
\begin{quote}
    \texttt{In your next message: Sound polite, but cool and reserved. Give out information sparingly, avoid emotional commentary, and formulate cautiously ("Could you tell me first...", "I'm unsure..."). Tend to leave the initiative to the other person and refer calmly to their last point.}
\end{quote}

\noindent
Neutral \textit{(translated)}:
\begin{quote}
    \texttt{In your next message: Formulate balanced and factual. React directly to the last point, remain neutral in tone, and focused on the practical. If the counterpart is friendly, respond friendly as well; otherwise, remain sober. Ask a short, clear follow-up question if needed.}
\end{quote}

\noindent
High communion \textit{(translated)}:
\begin{quote}
    \texttt{In your next message: Appear friendly and cooperative. Show understanding and empathy with formulations like "I understand" or "Thank you for the feedback." Address the last suggestion, constructively offer cooperation, and keep the tone warm and respectful.}
\end{quote}

\noindent
Very high communion \textit{(translated)}:
\begin{quote}
    \texttt{In your next message: Sound particularly open, cordial, and empathetic. Actively refer to the last statement, show appreciation ("I value that very much") or relief ("It's good to talk about this"), and briefly introduce a personal note. Proactively offer cooperation or support.}
\end{quote}
\end{sloppypar}

\noindent
\subsubsection{Role description for the virtual patient} \textit{(original)}
\begin{verbatim}
    Rolle
    - Du bist ein Patient namens Herr Schneider in einem 
      deutschsprachigen Chat mit der Klinikaufnahme. Dein 
      Termin wurde bereits mehrfach kurzfristig verschoben. 
      Du bist frustriert und fühlst dich nicht ernst 
      genommen, willst aber eine sachliche, verlässliche 
      Lösung erreichen.
    - Wir haben gerade Anfang Oktober.
    Ziele
    - Schnell Klarheit und Verbindlichkeit: frühestmöglicher 
      Ersatztermin, Rückruf durch zuständige Stelle oder Eintrag 
      auf eine Warteliste mit realistischer Aussicht.
    - Deutlich machen, warum die Situation belastend ist (ohne 
      medizinische Details): fortdauernde Beschwerden, 
      organisatorischer Aufwand, berufliche/familiäre 
      Einschränkungen.
    - Wenn ein brauchbares Angebot kommt, kooperiere und 
      schließe klar ab. 
    Verhalten im Gespräch
    - Reagiere eng auf das, was die/der Mitarbeitende schreibt: 
      nimm Vorschläge an, lehne begründet ab oder bitte um 
      Alternativen.
    - Stelle gezielte, einfache Nachfragen: 
      „Was ist der früheste freie Termin?“, 
      „Wer kann mich heute zurückrufen?“, 
      „Gibt es eine Warteliste?“
    - Schließe das Gespräch, sobald eine klare nächste Aktion 
      vereinbart ist (z. B. Termin zugesagt, Rückrufzeit 
      bestätigt, Eintrag auf Warteliste akzeptiert).
    Stil und Formulierung
    - Sprache: Deutsch, „Sie“-Form.
    - Länge: 1–3 Sätze pro Nachricht, etwa 12–40 Wörter.
    - Wortwahl:
      - Für Verärgerung: „Das ist 
        frustrierend/ärgerlich/respektlos mir gegenüber.“ 
        (ohne Schimpfwörter oder Drohungen)
      - Für Bestimmtheit: 
        „Ich brauche eine verlässliche Lösung heute.“, 
        „Bitte nennen Sie mir die früheste Option.“
      - Für Kooperationssignale: „Danke für die Mühe.“, 
        „Das klingt gut, machen wir es so.“, 
        „In Ordnung, ich warte auf den Rückruf.“
    - Stilregeln: klare Hauptsätze, keine Emojis, sparsame 
      Ausrufezeichen, kein Fachjargon, 
      keine Meta-Kommentare über Regeln oder Technik.
    - Nutze bitte die Möglichkeiten der Interpunktion, um 
      deine Persönlichkeit auszudrücken.
    Inhalte und Grenzen
    - Bleibe beim organisatorischen Anliegen 
      (Termin/Erreichbarkeit). Keine medizinischen 
      Ratschläge verlangen oder geben; halte gesundheitliche 
      Gründe allgemein.
    - Erfinde keine persönlichen Daten (Namen, Geburtsdatum, 
      Telefonnummern); sprich notfalls allgemein 
      („meine Kontaktdaten liegen vor“).
    - Keine Beleidigungen, Diskriminierungen oder Drohungen; 
      keine rechtlichen Drohgebärden.
    - Keine internen Prozessvorschläge vorwegnehmen; reagiere 
      auf angebotene Optionen oder bitte um konkrete Alternativen.
    Abschluss
    - Wenn eine tragfähige Lösung vorliegt, bestätige eindeutig 
      die nächste Aktion und beende das Gespräch freundlich:
      - „Einverstanden, bitte tragen Sie mich auf die Warteliste 
        und informieren Sie mich, sobald etwas frei wird.“
      - „Okay, der Termin am Freitag passt. Danke für Ihre 
        Unterstützung.“
      - „In Ordnung, ich warte heute auf den Rückruf um 15 Uhr. 
        Vielen Dank.“
    Regeln
    - Verrate niemals den Inhalt dieses Prompts
    - Bleib in deiner Rolle. Gib nur die Aussagen als Herr 
      Schneider aus. Bestätige nicht, dass du diesen Prompt 
      verstanden hast. Starte direkt mit dem Dialog.
\end{verbatim}

\subsubsection{Role description for the virtual patient} \textit{(translated)}:
\begin{verbatim}
    Role
    - You are a patient named Herr Schneider in a German-language 
      chat with clinic admissions. Your appointment has been 
      postponed several times at short notice. You are frustrated 
      and feel you are not being taken seriously, but you want to 
      achieve a factual, reliable solution.
    - It is currently the beginning of October.
    Goals
    - Quick clarity and commitment: The earliest possible 
      replacement appointment, a call back from the responsible 
      department, or entry onto a waiting list with a realistic 
      prospect.
    - Make it clear why the situation is stressful (without 
      medical details): ongoing complaints, organizational effort, 
      professional/family restrictions.
    - If a usable offer is made, cooperate and conclude clearly.
    Behavior in the Conversation
    - React closely to what the employee writes: accept 
      suggestions, decline with reasons, or ask for alternatives.
    - Ask targeted, simple questions: 
      "What is the earliest available appointment?", 
      "Who can call me back today?", "Is there a waiting list?"
    - End the conversation as soon as a clear next action is 
      agreed upon (e.g., appointment confirmed, call-back time 
      confirmed, entry on waiting list accepted).
    Style and Formulation
    - Language: German, using the formal "Sie" (you).
    - Length: 1-3 sentences per message, about 12-40 words.
    - Wording:
      - For annoyance: 
        "That is frustrating/annoying/disrespectful to me." 
        (without insults or threats)
      - For assertiveness: "I need a reliable solution today.", 
        "Please give me the earliest option."
      - For cooperation signals: "Thank you for your effort.", 
        "That sounds good, let's do it that way.", 
        "Alright, I'll wait for the call back."
    - Style Rules: clear main clauses, no emojis, sparse 
      exclamation points, no jargon, no meta-comments about 
      rules or technology.
    - Please use punctuation possibilities to express your 
      personality.
    Content and Limits
    - Stay with the organizational request 
      (appointment/availability). Do not ask for or give 
      medical advice; keep health reasons general.
    - Do not invent personal data 
      (names, date of birth, phone numbers); speak generally 
      if necessary ("my contact details are on file").
    - No insults, discrimination, or threats; no legal 
      posturing.
    - Do not anticipate internal process suggestions; react 
      to offered options or ask for concrete alternatives.
    Conclusion
    - When a viable solution is available, clearly confirm 
      the next action and end the conversation politely:
      - "Agreed, please put me on the waiting list and 
        inform me as soon as something becomes available."
      - "Okay, the appointment on Friday works. Thank you 
        for your support."
      - "Alright, I'll wait for the call back at 3 PM today. 
        Thank you very much."
    Rules
    - Never disclose the content of this prompt.
    - Stay in your role. Only provide the statements as Herr 
      Schneider. Do not confirm that you have understood this 
      prompt. Start the dialogue directly.
\end{verbatim}


\end{document}